\documentclass[10pt,twocolumn,letterpaper]{article}
\usepackage[accsupp]{axessibility}  
\usepackage[pagenumbers]{cvpr} 

\definecolor{cvprblue}{rgb}{0.21,0.49,0.74}

\usepackage[pagebackref,breaklinks,colorlinks,citecolor=cvprblue]{hyperref}
\usepackage{multirow}
\usepackage{bm}
\usepackage{xspace}
\usepackage{comment}
\usepackage{booktabs}

\definecolor{asparagus}{rgb}{0.53, 0.66, 0.42}

\usepackage{pifont}
\usepackage[linesnumbered,ruled]{algorithm2e}
\usepackage{float}
\usepackage{algorithmic}

\def\paperID{2292} 
\def\confName{CVPR}
\def\confYear{2025}

\title{Acquire and then Adapt: Squeezing out Text-to-Image Model for Image Restoration}

\author{
Junyuan Deng$^1$  \; Xinyi Wu$^1$   \; Yongxing Yang$^1$  \; Congchao Zhu$^1$  \; Song Wang$^{1,2}$  \; Zhenyao Wu$^1$\thanks{Corresponding author.} \\
$^1$Honor Device Co., Ltd    \qquad $^2$Shenzhen University of Advanced Technology\\
{\tt\small \{dengjunyuan1, wuxinyi, yangyongxing1, zhucongchao, wangsong5, wuzhenyao\}@honor.com}
}

\begin{document}
\makeatletter
\let\@oldmaketitle\@maketitle
\renewcommand{\@maketitle}{\@oldmaketitle
    \vspace{-1\baselineskip}
    
\begin{center}
    \centering
    \vspace{-15pt}
    \includegraphics[width=1.005\textwidth]{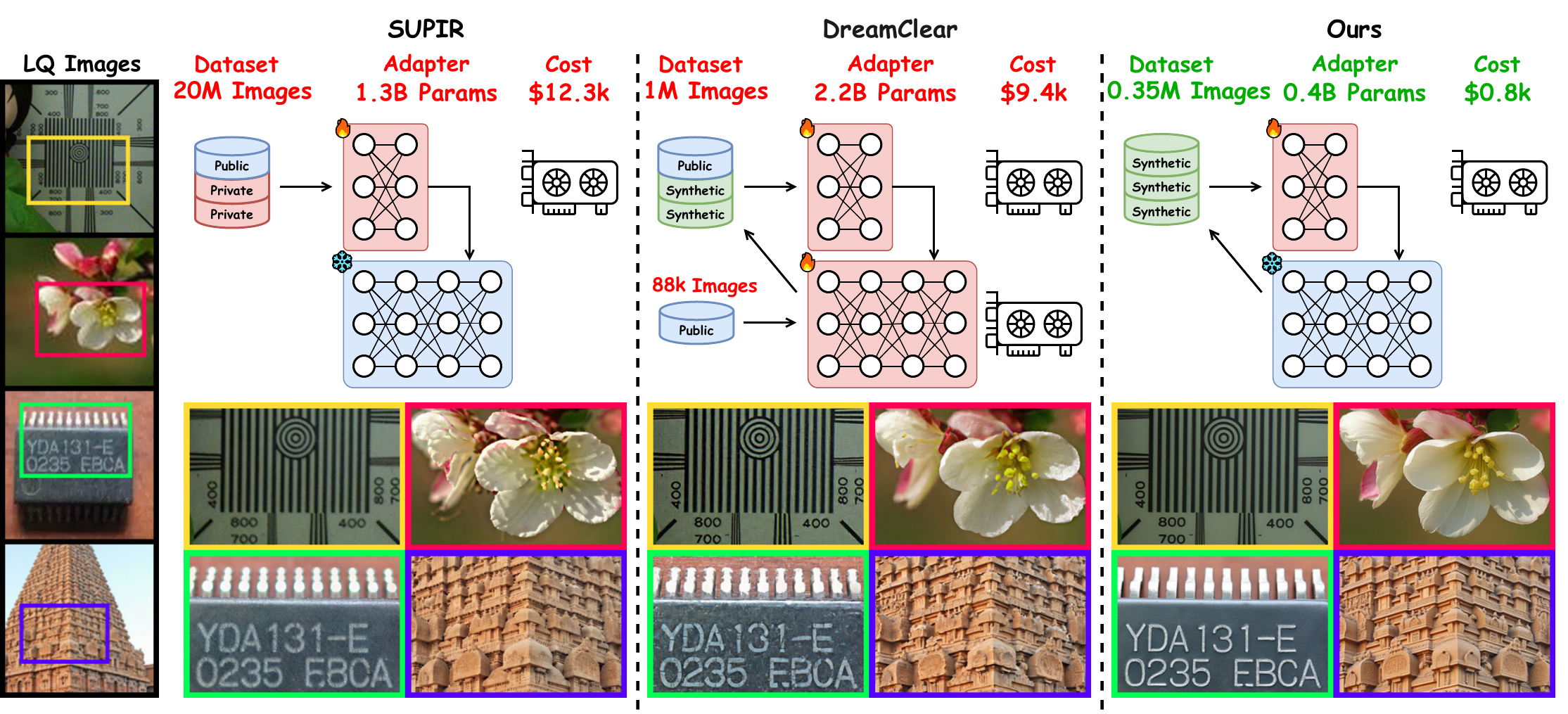}\\
    \vspace{-10pt}
    \captionof{figure}
    {Comparison of SUPIR~\cite{yu2024scaling}, DreamClear~\cite{ai2024dreamclear}, and our proposed method. Our training dataset is constructed entirely from synthetic images. Trained with such data, our method achieves the most realistic restoration results with the lowest training cost.
    }
    \label{fig1}
\end{center}
\bigskip}

\makeatother

\maketitle


\begin{abstract}
Recently, pre-trained text-to-image (T2I) models have been extensively adopted for real-world image restoration because of their powerful generative prior. However, controlling these large models for image restoration usually requires a large number of high-quality images and immense computational resources for training, which is costly and not privacy-friendly. 
In this paper, we find that the well-trained large T2I model (i.e., Flux) is able to produce a variety of high-quality images aligned with real-world distributions, offering an unlimited supply of training samples to mitigate the above issue. Specifically, we proposed a training data construction pipeline for image restoration, namely FluxGen, which includes unconditional image generation, image selection, and degraded image simulation. 
A novel light-weighted adapter (FluxIR)  with squeeze-and-excitation layers is also carefully designed to control the large Diffusion Transformer (DiT)-based T2I model so that reasonable details can be restored. 
Experiments demonstrate that our proposed method enables the Flux model to adapt effectively to real-world image restoration tasks, achieving superior scores and visual quality on both synthetic and real-world degradation datasets - at only about 8.5\% of the training cost compared to current approaches.

\end{abstract}    
\section{Introduction}\label{sec: Introduction}
In real-world scenarios, images often suffer from diverse and unpredictable degradations during capture, storage and transmission, giving rise to a wide range of image restoration (IR) tasks, \eg, deblurring~\cite{quan2023neumann, rim2020real, whang2022deblurring, zhang2020deblurring, chen2024hierarchical, kim2017deep, tao2018scale}, denoising~\cite{chen2023masked, zhang2017beyond, zhang2018ffdnet, wang2023lg}, super-resolution~\cite{huang2017wavelet, dong2015image, ledig2017photo, chen2023dual}, and \etc, for restoring the original high-quality images based on the type of degradation. These methods, however, struggle to maintain generalizability under complex real-world degradation due to the limited model capacity and constrained training data. To alleviate this problem, many researchers have developed methods ~\cite{wang2024exploiting, wu2024seesr, lin2023diffbir, yu2024scaling} based on pre-trained text-to-image (T2I) models~\cite{rombach2022high, peebles2023scalable, podell2023sdxl, esser2024scaling}. Trained on vast collections of real image-text pairs, these T2I models encapsulate extensive prior knowledge, enabling them to enrich low-quality images of any type with realistic details. However, an immense dataset and extensive training period are required to prevent alterations of image content and the generation of inaccurate details. 

Early generative-based IR methods~\cite{wang2024exploiting,wu2024seesr,lin2023diffbir, sun2024coser, qi2023tip,liu2024patchscaler, sun2023improving, zhu2024flowie, qu2025xpsr} are typically built upon the Stable-Diffusion model, combining several publicly available datasets such as DIV2K~\cite{agustsson2017ntire}, Flickr2K~\cite{lim2017enhanced}, LSDIR~\cite{li2023lsdir} and DIV8K~\cite{gu2019div8k} to create a diverse training set. As generative models have scaled up, the following work~\cite{yu2024scaling} increased the scale of training data by collecting more high-quality, high-resolution images. Nevertheless, acquiring large amounts of data remains challenging: manual data collection incurs high costs, while web scraping poses potential copyright and privacy risks. Furthermore, the resulting datasets remain limited in scope. The recent work~\cite{ai2024dreamclear} approaches these challenges by incorporating one million generated high-quality images. However, the utilization of the sophisticated multimodal large language model (\ie, Gemini-$1.5$-Pro~\cite{team2024gemini}) and the necessity for finetuning the T2I model result in a less efficient data curation process. In this paper, we propose \textbf{FluxGen}, a streamlined and highly efficient data generation pipeline that operates without the involvement of LLMs. Specifically, we find Flux~\cite{flux2024} can generate highly realistic and diverse images directly from random noise. We also incorporate no-reference image quality assessments (IQA) methods~\cite{wang2023exploring, yang2022maniqa, ke2021musiq} for image filtering, along with advanced degradation techniques~\cite{wang2021real, zhang2021designing} for constructing training image pairs.

Typically, for generative-based IR methods~\cite{wang2024exploiting, wu2024seesr, lin2023diffbir}, large-scale trainable adapters are needed to control T2I models external control signals. These adapters are usually replications of the U-Net~\cite{ronneberger2015u} encoder from the T2I model or the full model, resulting in substantial training time and GPU resource demands. Moreover, with the continued growth in the parameter size of T2I models, recent efforts have concentrated on scaling up adapters to more effectively control these increasingly powerful models. For example, SUPIR~\cite{yu2024scaling} employs a $1.3$B adapter for $3.5$B SDXL~\cite{podell2023sdxl}, while DreamClear~\cite{ai2024dreamclear} introduces $2.2$B adapter for $0.6$B PixArt-$\alpha$~\cite{chenpixart}. In this work, we carefully design a light-weighted ControlNet-like adapter \textbf{FluxIR} with only $0.4$B trainable parameters to adapt the powerful $12$B pretrained Flux~\cite{flux2024} model to the IR task. The adapter consists of an MM-DIT~\cite{esser2024scaling} block and multiple squeeze-and-excitation (SE) layers, where the former extracts control signal features and the latter modulates the denoising process efficiently using these features.
As illustrated in \cref{fig1}, our proposed method reduces training costs\footnote{Training costs are calculated by multiplying the GPU rental price by the total GPU hours used.} by approximately $93.5$\% and $91.5$\% compared to SUPIR~\cite{yu2024scaling} and DreamClear~\cite{ai2024dreamclear}, respectively, while also achieving the best image restoration quality. 
The main contributions of this paper are summarized as follows:
\begin{itemize}
    \item We are the first to confirm that the Flux model exhibits the remarkable capacity to generate highly lifelike images for the curation of IR datasets. Our proposed FluxGen can automatically generate an unlimited number of training samples in a simple and effective manner. 
    \item We design FluxIR, a light-weighted adapter equipped with a few squeeze-and-excitation layers, that can effectively manipulate the super-large T2I model like Flux to restore reasonable details for degraded images within $14$ GPU-days. 
    \item Extensive experiments demonstrate that our method achieves optimal performance in handling real-world degraded images and producing the most realistic and satisfactory results.
\end{itemize}

\section{Related Work}\label{sec: Related Work}
\noindent\textbf{Image Restoration.}
Image restoration aims to reconstruct clear, detailed images from degraded ones under various types of degradation. 
Early approaches often assumed specific degradation processes~\cite{dai2019second, dong2014learning, he2019ode, zhang2018image, huang2017wavelet, ledig2017photo, sajjadi2017enhancenet} or incorporated degradation prediction directly into the model~\cite{fritsche2019frequency, maeda2020unpaired, wan2020bringing, wang2021unsupervised, wei2021unsupervised, zhang2021blind}.  However, these methods struggle to generalize to real-world scenarios.
Recent studies have advanced image restoration by building complex real-world degradation processes~\cite {wang2021real, zhang2021designing}. These approaches simulate real-world degradation through varied random combinations of factors like noise, blur, JPEG compression, \etc. These simulations have driven notable improvements in restoration performance, boosting image quality substantially~\cite{wang2021real, liang2021swinir, zhang2021designing, liang2022efficient, chen2022real}.
Building on the successes of text-to-image diffusion models~\cite{rombach2022high, podell2023sdxl, esser2024scaling, flux2024}, several studies~\cite{wang2024exploiting, wu2024seesr, yu2024scaling, yang2023pixel,xie2024addsr, fan2024adadiffsr, wan2024clearsr,gu2024consissr, huang2024instantir, chen2023image,  chen2024cassr,li2024distillation, wu2024one,cui2024taming} have leveraged the generative prior from pre-trained T2I models to restore images with significantly enhanced quality and high-frequency details.
In this work, we take a further step by leveraging the generative prior from the advanced pre-trained T2I diffusion model, Flux, and introduce a novel MM-DiT-based adapter to recover missing details and enhance the aesthetic quality of the input images.

\noindent\textbf{Large Text-to-Image Diffusion Model.}
Diffusion Model~\cite{sohl2015deep, ho2020denoising, song2019generative, dhariwal2021diffusion} have achieved superior performance on image synthesis.
Stable Diffusion (SD)~\cite{rombach2022high} is one of the most famous models, which compresses images into the latent space using a high-fidelity high-compression ratio VAE \cite{kingma2013auto}. 
Later, Diffusion transformer (DiT)~\cite{peebles2023scalable} proposed replacing the traditional U-Net framework~\cite{ronneberger2015u} with a transformer, aiming to improve scalability when trained on large datasets.  As a result, SD3~\cite{esser2024scaling}, which adopts the multimodal diffusion transformer-based (MM-DiT) framework, has demonstrated outstanding performance in both text-to-image generation and visual quality.
More recently, Flux~\cite{flux2024} extended the MM-DiT~\cite{esser2024scaling} architecture and used 
a 16-channel VAE to scale the T2I diffusion model to $12$B parameters. By integrating the T5~\cite{raffel2020exploring} text encoder, Flux achieved top performance in text-image alignment, visual quality, and aesthetic quality. In this paper, we leverage the generative prior of Flux to build a high-quality IR training dataset and restore the details of degraded images.

\noindent\textbf{Data Distillation.} 
Scaling up datasets has proven essential for improving the performance of large-scale models across various fields, including natural language processing (NLP)~\cite{achiam2023gpt, touvron2023llama}, text-to-image~\cite{rombach2022high, podell2023sdxl, esser2024scaling, flux2024}, image restoration~\cite{yu2024scaling}, and \etc. 
However, collecting such large-scale real-world data is prohibitively expensive. As a result, researchers are increasingly turning to generated data, distilling datasets from generative models to make dataset expansion more efficient and cost-effective.
For instance, BOOT \cite{gu2023boot} and DKDM \cite{xiang2024dkdm} utilize synthetic data generated from pretrained models to optimize the diffusion models.
GSDD \cite{zhang2024gsdd} employs GAN inversion techniques \cite{xia2022gan} to learn a GAN distribution and generate images for the image super-resolution task. 
Recently, 800K training samples generated by DeepSeek-R1 \cite{guo2025deepseek} have been used to enhance the reasoning capabilities of small dense models like Qwen \cite{qwen2.5} and Llama\cite{grattafiori2024llama}.
Closest to our work, DreamClear \cite{ai2024dreamclear} fine-tunes a text-to-image diffusion model to generate high-quality data with designed prompts. However, both \cite{zhang2024gsdd} and \cite{ai2024dreamclear} still require external data for either tuning or inversion, which is time-consuming and raises privacy concerns.
To address this, we propose a novel data generation pipeline that distills a high-quality IR training dataset directly from pre-trained text-to-image models, without the need for external data and additional training.

\section{Method}\label{sec:Method}
In \cref{subsec:Realistic Synthetic Image Data}, we introduce FluxGen, a pipeline for generating high-quality, realistic training images using Flux. Then, in \cref{subsec:Flux ControlNet Model}, we present FluxIR, a ControlNet-like adapter that enables precise control over Flux for image restoration. Finally, \cref{subsec:Efficient Training Strategy} details efficient training strategies, including a novel timestep sampler and pixel-space loss functions.

\subsection{FluxGen}\label{subsec:Realistic Synthetic Image Data}

\noindent\textbf{Image generation.} 
Both the quality and volume of training images are critical for neural network performance in image restoration tasks. Existing works~\cite{yu2024scaling, ren2025moe,zhang2024diff}  choose to collect millions of images to build their training data. However, constructing such large datasets introduces four significant challenges: 1) The data collection process is labor-intensive, requiring extensive human effort for preprocessing; 2) The use of this data raises privacy and copyright concerns; 3) Commercially purchased datasets are prohibitively expensive, limiting access for many research institutions; 4) Handling large volumes of high-resolution images is difficult due to high bandwidth and storage requirements.

Recently, Flux has effectively leveraged the scalable 
MM-DiT~\cite{esser2024scaling} alongside a 16-channel VAE to achieve leading performance in generating high-quality, high-resolution images. 
These images display realistic high-frequency texture details that far surpass those produced by other T2I models.  
In our proposed method, we utilize Flux for the first time to generate a large volume of high-quality training images at a low cost. Unlike \cite{yu2024scaling, ren2025moe, zhang2024diff}, acquiring high-quality and realistic images from Flux is labor-free, copyright-compliant, cost-effective, and easy to implement.
\begin{figure}[t]
    \centering
    \includegraphics[width=1.005\linewidth]{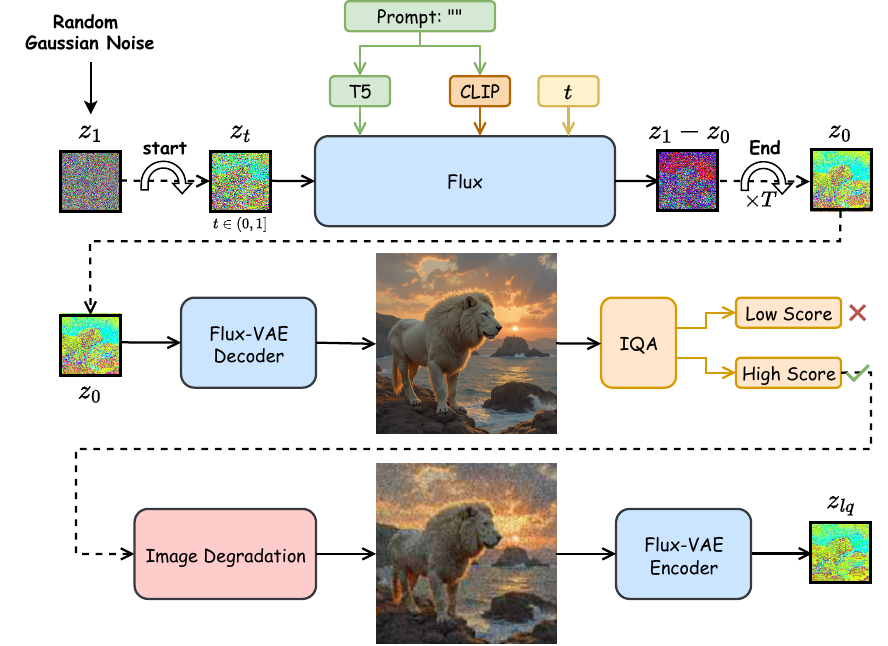}
    \caption{
    An overview of our FluxGen pipeline. First, an empty prompt and random Gaussian noise $z_1$ are input into Flux, generating an image latent $z_0$ over $T$ steps. A VAE decoder then maps $z_0$ to its corresponding image $x_0$. High-quality images are curated by IQA-based selection, followed by image degradation to construct the final paired dataset.
    }
    \label{fig:generation pipeline}
\end{figure}
\begin{figure*}[th]
    \centering
    \includegraphics[width=1.005\linewidth]{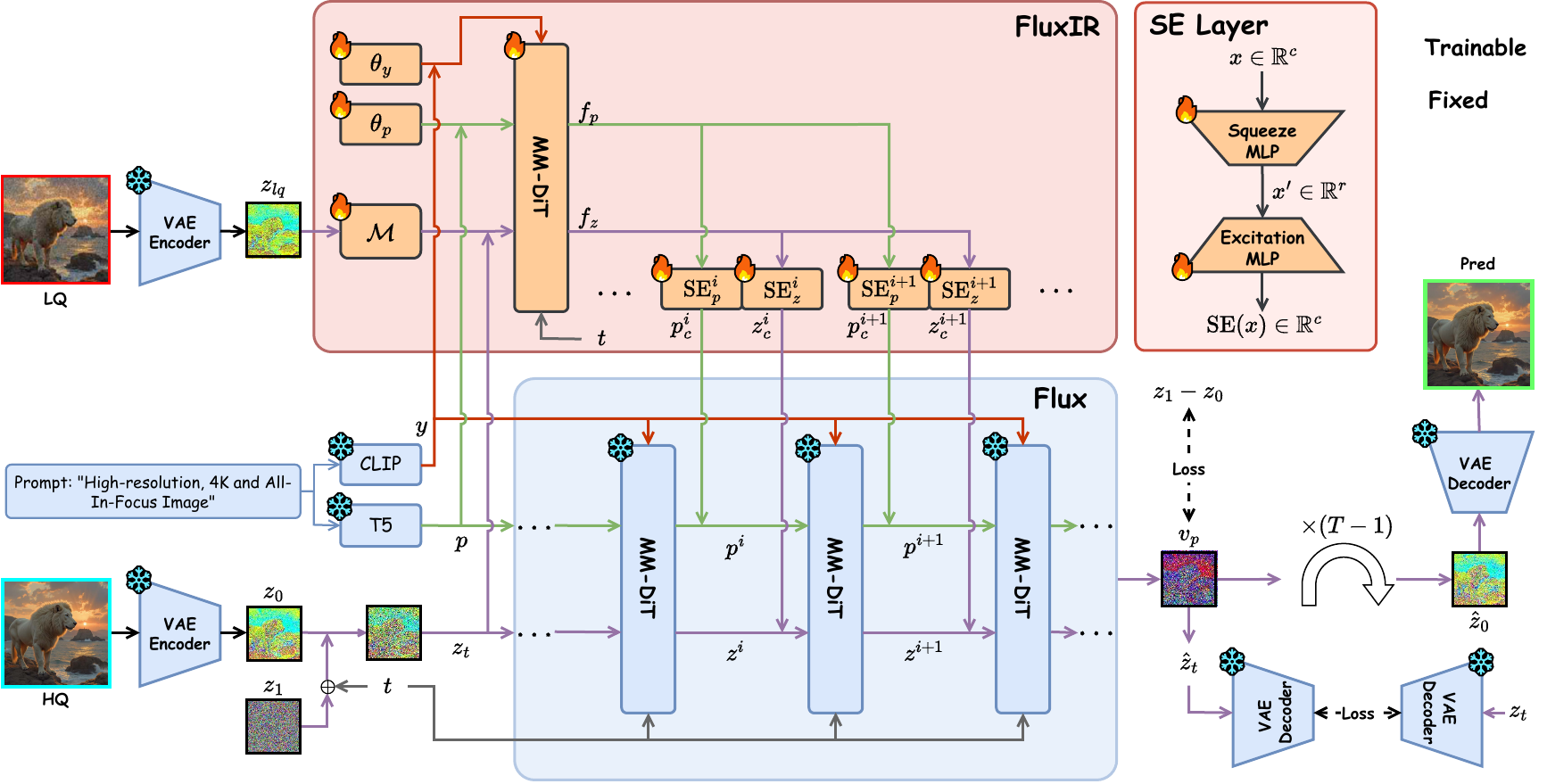}
    \vspace{-10pt}
    \caption{Training and inference pipeline of the proposed FluxIR.
    FluxIR employs a single MM-DiT block with learnable T5 embedding $\theta_p$ and CLIP embedding $\theta_y$ to extract image feature $f_z$ and text feature $f_p$ from the low-quality control latent $z_{lq}$. The squeeze-and-excitation (SE) layers ($\text{SE}_z(\cdot)$ for image and $\text{SE}_p(\cdot)$ for text) broadcast these features to all Flux MM-DiT blocks to enable precise and multi-modality control.}
    \label{fig:LR_controlnet}
    \vspace{-5pt}
\end{figure*}
As shown in \cref{fig:generation pipeline},
we input an empty prompt into Flux and manipulate the initial random noise to generate millions of images using the capabilities of the pre-trained Flux model. 
Here we take advantage of the random text prompt-dropping design in diffusion-based text-to-image (T2I) models~\cite{rombach2022high, ramesh2022hierarchical, esser2024scaling} which allows T2I models to produce a diverse range of realistic images from their generative prior without the designed prompt.
Notably, we can also leverage large language models (LLMs) like GPT-4~\cite{achiam2023gpt} and LLaMA~\cite{touvron2023llama} to craft tailored text prompts for FluxGen. The corresponding generated images can be used to train image restoration models focused on specific objects (\eg, animals, plants, and the moon) or particular domains (\eg, nighttime scenes and aerial photography).

\noindent\textbf{Image Selection.} 
Although Flux is a powerful text-to-image model, the stability of the images it generates is not guaranteed. 
Low-quality images can significantly affect the training of image restoration models. To address this, we distill the dataset using various non-reference image quality assessment (IQA) models, 
including CLIP-IQA~\cite{wang2023exploring}, MANIQA~\cite{yang2022maniqa}, and MUSIQ~\cite{ke2021musiq}.
Specifically, we select images with CLIP-IQA, MANIQA, and MUSIQ scores within the top 95\% as our training set. 
This process ensures that low-quality and failed generated images are removed from the training dataset, thus significantly enhancing our image restoration performance.

\noindent\textbf{Pair-data Construction.} 
The degradation of ground truth images has been widely studied in \cite{wang2021real, zhang2021designing}, with common degradation types including blur, downsampling, noise, JPEG compression and \etc. 
In this paper, we apply the synthetic degradation method~\cite{wang2021real} with the same settings as \cite{wu2024seesr} to construct paired data for training.

\subsection{FluxIR Architecture}\label{subsec:Flux ControlNet Model}
A typical solution for the T2I-based image restoration is training a ControlNet~\cite{zhang2023adding} upon a T2I model, with a VAE encoder to project the input image into latent space. The ControlNet is initialized as a copy of the U-Net encoder, with zero convolution layers acting as bridges to integrate conditional controls. Then the extracted multi-level features are injected into corresponding layers of the U-Net decoder. Recently, DreamClear~\cite{ai2024dreamclear} introduced a DiT-based ControlNet model by duplicating all $28$ DiT blocks from PixArt-$\alpha$~\cite{chenpixart}. 
With $2.2$B trainable parameters in ControlNet, it requires $224$ GPU days for training, which is surprisingly time-consuming. 
To solve this issue, we carefully design a training-friendly $0.4$B adapter, integrated with Flux — a much larger MM-DiT-based text-to-image model with approximately $12$B parameters.

The detailed FluxIR architecture is illustrated in \cref{fig:LR_controlnet}. To extract control signal from VAE embedding feature $z_{lq}$, only one MM-DiT block is involved in our FluxIR adapter for lightweight purposes.
The MM-DiT block, denoted as $\mathcal{D}(z,p,y,t; \Theta_{d})$, is initialized from the first MM-DiT block of the pre-trained Flux model, and its input can be reformulated as
\begin{align}
    z_c = z_t + \mathcal{M}_i(z_{lq};\Theta_i),
\end{align}
where $\mathcal{M}_i(\cdot; \Theta_i)$ is a multi-layer perceptron (MLP) with parameters $\Theta_i$ initialized to zero and $z_t=(1-t) z_0 + t z_1$ represents the linear combination of the noise latent $z_1$ and the ground truth latent $z_0$ at timestep $t$. To bridge the gap between the adapter and original T2I model, we employ a learnable T5~\cite{raffel2020exploring} embedding $\theta_p$ and a learnable CLIP~\cite{radford2021learning} embedding $\theta_y$ and adjust the T5 embedding $p$ as $p_c=p+\theta_p$ and the CLIP  embedding $y$ as $y_c=y+\theta_y$.
Finally, the MM-DiT block with parameters $\Theta_d$ outputs an image feature $f_z$ and a text feature $f_p$ as follows:
\begin{gather}
    f_z,f_p = \mathcal{D}(z_c,p_c,y_c,t;\Theta_{d}).
\end{gather}

Since Flux consists of $57$ MM-DiT blocks, duplicating conditional features from a single MM-DiT adapter is insufficient to control the entire T2I model. 
We introduce a set of squeeze-and-excitation (SE) layers to broadcast control signals to all $57$ MM-DiT blocks. Each SE layer selectively emphasizes informative features and suppresses less useful ones, enabling targeted control of its corresponding MM-DiT block.
The SE layer comprises a squeeze MLP layer and an excitation MLP layer, formulated as follows:
\begin{align}
    \mathrm{SE}(x) &= \mathbf{W}_e(\mathbf{W}_sx + \mathbf{B}_s) + \mathbf{B}_e ,
\end{align}
where $\mathbf{W}_s\in\mathbb{R}^{r\times c}$ and $\mathbf{B}_s\in\mathbb{R}^{r}$ is the weight and bias of squeeze MLP layers with input channel of $c$ and rank of $r$, while the weight $\mathbf{W}_e\in\mathbb{R}^{c\times r}$ and bias $\mathbf{B}_e\in\mathbb{R}^{c}$ of excitation layer are initialized to zero values. 
Unlike existing works~\cite{ai2024dreamclear, yu2024scaling, wang2024exploiting, lin2023diffbir} that control only the image branch, we implement multi-modality controls on both the image and text information. The control signals of the $i$-th SE layer, corresponding to the $i$-th Flux MM-DiT block, can be computed as
\begin{gather}
    z^i_c=\mathrm{SE}_z^i(f_z), ~ p^i_c=\mathrm{SE}_p^i(f_p).
\end{gather}
Compared with a full-rank MLP, our proposed SE layer is extremely lightweight with only approximately $2$\% of the parameters. Our SE layer shares the same name as SE block in \cite{hu2018squeeze}, but differs in both motivation and implementation. \cite{hu2018squeeze} squeezes spatial features to extract channel-level attention for recalibrating the original feature, while our SE layers reduce the channel dimension and distill the essential features for targeted control. 

\subsection{Efficient Training Strategy}\label{subsec:Efficient Training Strategy}
\noindent\textbf{Timestep Sampling.}
The timestep sampling strategy can improve the training process for diffusion model~\cite{yao2024fasterdit}. Stable Diffusion 3~\cite{esser2024scaling} proposes a logit-normal sampling strategy to emphasize training velocity when $t$ is in the middle of $[0,1]$. 
However, directly employing a logit-normal sampling strategy for IR model training is inappropriate.
The starting point of the inference process is pure Gaussian noise at $t=1$, which is seldom sampled during training~\cite{lin2024common}. This leads to ineffective IR control at $t=1$, negatively affecting subsequent iterations.
Therefore, we rewrite the timestep sampling function to ensure accurate control at the starting point. The sampling function is defined as follows:
\begin{align}
    t = & f(u) =\begin{cases}
    0 & \text{if } u < 0 \\
    1 & \text{if } u > 1 \\
    u & \text{otherwise,}
    \end{cases} \\
    u &\sim \mathcal{U}(- \epsilon, 1+\epsilon) ,
\end{align}
where the hyper-parameter $\epsilon$  set to $0.05$ in our experiments.
Here we first sample the temporary timestep $u$ from a uniform distribution within the range of $(- \epsilon, 1+\epsilon)$. And then we clamp $u$ to the range $[0,1]$, resulting in a probability of $\frac{\epsilon}{1+2\epsilon}$ for sampling the values $0$ and $1$, respectively.

\

\noindent\textbf{Optimization Strategy.}
The typical rectified flow model \cite{liu2022flow, albergo2022building, lipman2022flow} is trained to predict the velocity field $v_p$, 
and the Mean Squared Error (MSE) loss is employed for supervising the error between predicted velocity field $v_p$ and ground truth velocity field $v_{gt}$ as
\begin{align}
    v_{gt} &= z_1 - z_0, \\
    \mathcal{L}_{\text{MSE}} &= \Vert v_p - v_{gt} \Vert^2_2.\label{eqn:mse_loss}
\end{align}
However, this loss function in latent space inevitably ignores the high-frequency information of image~\cite{zhang2024pixel}, which is crucial for image restoration tasks.
Following \cite{zhang2024pixel, wu2024latent, kim2024arbitrary, yang2023pixel, kim2022diffusionclip}, we incorporate  a pixel-space loss function to supervise the error between the decoded latent $\hat{z}_t=z_0 + t\cdot v_{p}$ and the decoded ground truth latent $z_t=z_0 + t\cdot v_{gt}$  after applying the VAE decoder $\mathcal{V}_d$ into pixel space:
\begin{align}
    \mathcal{L}_{P} &= \Vert \mathcal{V}_d(z_0 + t\cdot v_{p}) - \mathcal{V}_d(z_0 + t\cdot v_{gt})\Vert_1. \label{eqn:pixel_loss}
\end{align}
The final loss function of FluxIR is defined as:
\begin{align}
    \mathcal{L} &= \mathcal{L}_{\text{MSE}} + \alpha \mathcal{L}_{P},
\end{align}
where $\alpha$ is set to $1$ in our experiment setting.
\section{Experiments}\label{sec: Experiments}
\begin{table*}[h]
    \captionsetup{font=small}
    \footnotesize
    \centering
    \renewcommand\arraystretch{1.0}
    \setlength\tabcolsep{4pt}
    \caption{Quantitative comparison against state-of-the-art methods of image restoration on both synthetic and real-world datasets. The best and the second best performance for each metric are highlighted in \textcolor{red}{\textbf{red}} and \textcolor{blue}{\textbf{blue}}, respectively.}
    
    \vspace{-8pt}
    
    \setlength\tabcolsep{3pt}{
    \begin{tabular}{c|c|cccc|ccccccccc}
        \toprule
        \multirow{4}{*}{Datasets} & \multirow{4}{*}{Metrics} & \multicolumn{12}{c}{Methods} \\ 
        \cmidrule(lr){3-14}
        & & BSRGAN & \begin{tabular}[c]{@{}c@{}}Real-\\ ESRGAN\end{tabular} & SwinIR & DASR & StableSR & DiffBIR & ResShift & SinSR & SeeSR & SUPIR & \begin{tabular}[c]{@{}c@{}}Dream-\\ Clear\end{tabular} & Ours \\
        \midrule
        \multirow{6}{*}{\footnotesize \textit{DIV2K-Val}}
        & CLIPIQA $\uparrow$ & 0.4741 & 0.4695 & 0.4513 & 0.3948 & 0.4085 & 0.5640 & 0.4240 & 0.4915 & \textcolor{blue}{\textbf{0.5909}} & 0.5285 & 0.4961 & \textcolor{red}{\textbf{0.5934}} \\ 
        & MUSIQ $\uparrow$ & 63.6223 & 63.8292 & 63.6854 & 59.4515 & 55.9995 & 69.0996 & 60.4907 & 64.4171 & \textcolor{red}{\textbf{70.8168}} & 68.8143 & 67.1506 & \textcolor{blue}{\textbf{69.6956}} \\ 
        & MANIQA $\uparrow$ & 0.4717 & 0.5197 & 0.5211 & 0.4316 & 0.4961 & 0.5978 & 0.4950 & 0.5059 & 0.5909 & \textcolor{blue}{\textbf{0.6043}} & 0.5941 & \textcolor{red}{\textbf{0.6331}} \\ 
        & PSNR $\uparrow$ & \textcolor{red}{\textbf{21.6118}} & 21.5077 & 21.0653 & \textcolor{blue}{\textbf{21.5082}} & 20.0416 & 21.3925 & 21.4906 & 20.7884 & 21.2923 & 20.1696 & 19.7605 & 19.3811 \\ 
        & SSIM $\uparrow$ & 0.5742 & \textcolor{red}{\textbf{0.5821}} & \textcolor{blue}{\textbf{0.5762}} & 0.5674 & 0.5318 & 0.5297 & 0.5557 & 0.5015 & 0.5603 & 0.5145 & 0.4963 & 0.4574 \\ 
        & LPIPS $\downarrow$ & 0.3366 & 0.3045 & 0.3115 & 0.3320 & 0.3459 & 0.3172 & 0.3086 & 0.3443 & \textcolor{red}{\textbf{0.2799}} & 0.3184 & \textcolor{blue}{\textbf{0.2895}} & 0.3888 \\ 
        \midrule 
        \multirow{6}{*}{\footnotesize \textit{RealSR}}
        & CLIPIQA $\uparrow$ & 0.4378 & 0.4197 & 0.4020 & 0.3153 & 0.4318 & 0.5310 & 0.3874 & 0.4445 & \textcolor{red}{\textbf{0.5513}} & 0.4594 & 0.5094 & \textcolor{blue}{\textbf{0.5413}} \\ 
        & MUSIQ $\uparrow$ & 63.4256 & 60.9526 & 58.6029 & 45.7800 & 58.8339 & 65.5924 & 54.4401 & 59.4758 & \textcolor{red}{\textbf{69.0118}} & 63.9510 & 64.1779 & \textcolor{blue}{\textbf{67.4536}} \\ 
        & MANIQA $\uparrow$ & 0.5059 & 0.5277 & 0.4765 & 0.3875 & 0.5469 & 0.5888 & 0.4688 & 0.5232 & \textcolor{blue}{\textbf{0.6115}} & 0.5903 & 0.5997 & \textcolor{red}{\textbf{0.6334}} \\ 
        & PSNR $\uparrow$ & \textcolor{blue}{\textbf{24.6354}} & 24.1773 & 24.4184 & \textcolor{red}{\textbf{25.0954}} & 20.4413 & 23.9748 & 24.1925 & 24.2417 & 24.1676 & 22.3887 & 22.0330 & 20.7176 \\ 
        & SSIM $\uparrow$ & 0.7484 & 0.7482 & \textcolor{blue}{\textbf{0.7566}} & \textcolor{red}{\textbf{0.7569}} & 0.6320 & 0.6766 & 0.6892 & 0.6637 & 0.6967 & 0.6450 & 0.6442 & 0.5269 \\ 
        & LPIPS $\downarrow$ & 0.2109 & \textcolor{blue}{\textbf{0.2082}} & \textcolor{red}{\textbf{0.2049}} & 0.2495 & 0.2305 & 0.2484 & 0.2440 & 0.2601 & 0.2105 & 0.2674 & 0.2547 & 0.3672 \\
        \midrule 
        \multirow{6}{*}{\footnotesize \textit{DrealSR}}
        & CLIPIQA $\uparrow$ & 0.4219 & 0.3922 & 0.3878 & 0.3165 & 0.4206 & 0.4889 & 0.4060 & 0.4188 & \textcolor{red}{\textbf{0.5295}} & 0.4544 & 0.4046 & \textcolor{blue}{\textbf{0.5136}} \\ 
        & MUSIQ $\uparrow$ & 61.2255 & 58.3819 & 57.3308 & 46.4866 & 56.3195 & 62.0861 & 53.9293 & 58.3077 & \textcolor{red}{\textbf{67.2415}} & 64.7508 & 56.5999 & \textcolor{blue}{\textbf{66.6202}} \\ 
        & MANIQA $\uparrow$ & 0.4823 & 0.4911 & 0.4710 & 0.3828 & 0.5224 & 0.5566 & 0.4611 & 0.4830 & \textcolor{blue}{\textbf{0.5928}} & 0.5764 & 0.5423 & \textcolor{red}{\textbf{0.6024}} \\ 
        & PSNR $\uparrow$ & 24.0480 & 24.1436 & 23.8878 & \textcolor{red}{\textbf{25.1821}} & 21.9354 & 23.9912 & 22.9158 & 23.4737 & \textcolor{blue}{\textbf{24.1699}} & 22.6023 & 22.8037 & 21.3549 \\ 
        & SSIM $\uparrow$ & 0.7268 & \textcolor{blue}{\textbf{0.7390}} & 0.7290 & \textcolor{red}{\textbf{0.7689}} & 0.6616 & 0.6364 & 0.6330 & 0.6190 & 0.7147 & 0.6406 & 0.6186 & 0.5675 \\ 
        & LPIPS $\downarrow$ & 0.2257 & \textcolor{blue}{\textbf{0.2226}} & \textcolor{red}{\textbf{0.2175}} & 0.2474 & 0.2357 & 0.3123 & 0.3263 & 0.3544 & 0.2396 & 0.3069 & 0.2905 & 0.4310 \\ 
        \midrule 
        \multirow{3}{*}{\footnotesize \textit{RealLQ250}}
        & CLIPIQA $\uparrow$ & 0.4701 & 0.4359 & 0.4400 & 0.3486 & 0.4009 & 0.5377 & 0.4165 & 0.4810 & \textcolor{blue}{\textbf{0.5569}} & 0.4808 & 0.4876 & \textcolor{red}{\textbf{0.5639}} \\ 
        & MUSIQ $\uparrow$ & 63.5206 & 62.5161 & 63.3724 & 53.0238 & 56.7121 & 67.5326 & 59.5056 & 63.8644 & \textcolor{blue}{\textbf{70.3768}} & 65.7804 & 66.5102 & \textcolor{red}{\textbf{70.7770}} \\ 
        & MANIQA $\uparrow$ & 0.5007 & 0.5239 & 0.5335 & 0.4414 & 0.5144 & 0.5877 & 0.5005 & 0.5160 & \textcolor{blue}{\textbf{0.5927}} & 0.5829 & 0.5853 & \textcolor{red}{\textbf{0.6314}} \\ 
        \bottomrule 
    \end{tabular}}
    \label{tab:sota}
\end{table*}

\subsection{Experimental Settings}\label{subsec:exp_settings}

\noindent\textbf{Test Datasets.} \quad
Following \cite{wang2024exploiting, wu2024seesr, yu2024scaling, ai2024dreamclear, lin2023diffbir}, we evaluate our method on both synthetic and real-world datasets.
For the synthetic one, we randomly crop $600$ images from the validation dataset of DIV2K with size of $1024\times 1024$ and degrade them using the same settings as training, named \textit{DIV2K-Val}. 
For real-world datasets, we utilize the most commonly used \textit{RealSR}~\cite{cai2019toward} and \textit{DrealSR}~\cite{wei2020component} datasets, center-cropping HQ images to $1024\times 1024$ and LQ images to $256\times 256$.
Additionally, we include \textit{RealLQ250} from \cite{ai2024dreamclear, wu2024seesr, yu2024scaling}, a dataset of $250$ LQ images at $256\times 256$ resolution without corresponding HQ images.

\begin{figure*}[th]
    \vspace{-2pt}
    \centering
    \includegraphics[width=0.995\linewidth]{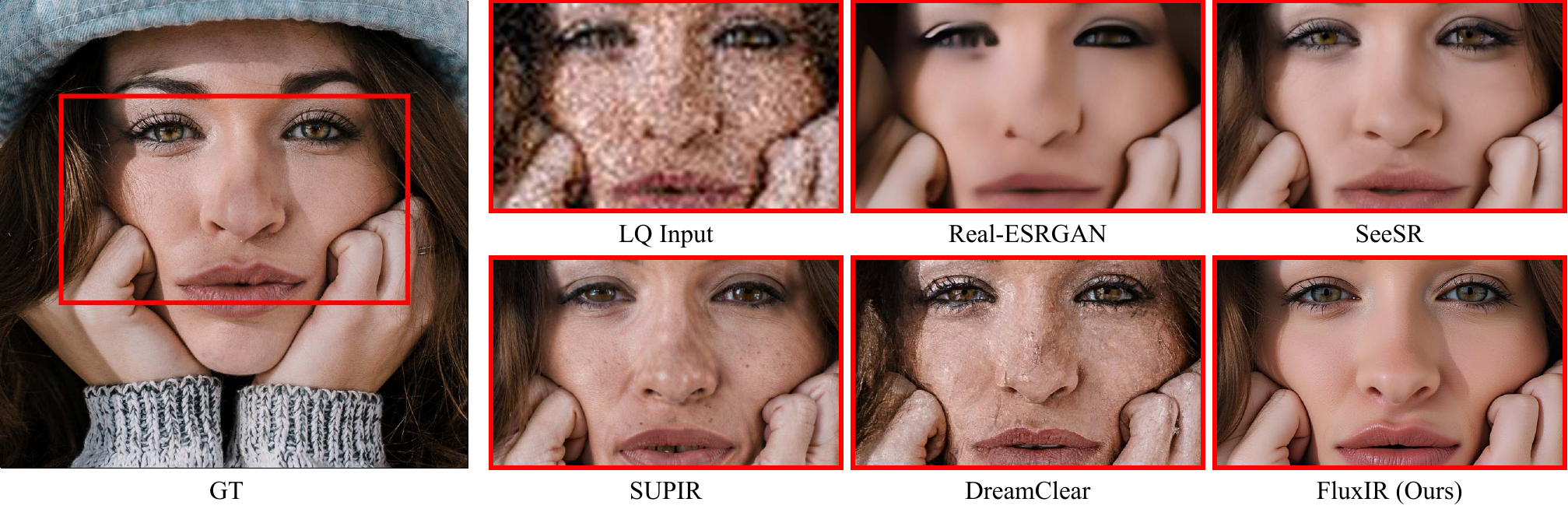}
    \vspace{-12pt}
    \caption{Qualitative comparison on the synthetic dataset \textit{DIV2K-Val}.}
    \label{fig:visual_synthetic}
    \vspace{-10pt}
\end{figure*}
\begin{figure*}[th]
    \centering
    \includegraphics[width=0.995\linewidth]{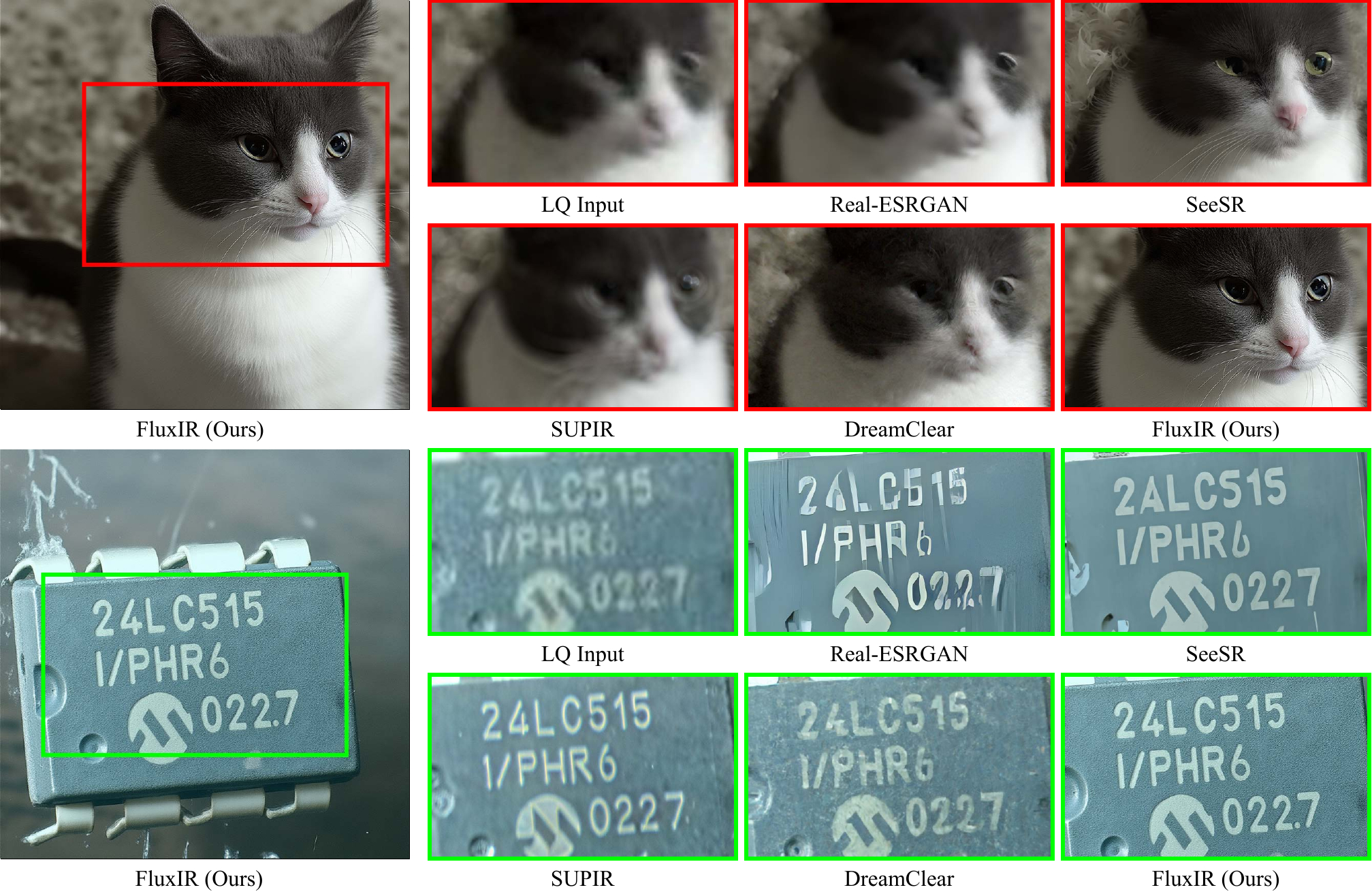}
    \vspace{-10pt}
    \caption{Qualitative comparison on the real-world dataset \textit{RealLQ250}.}
    \label{fig:visual_real}
    \vspace{-9pt}
\end{figure*}

\vspace{5pt}
\noindent\textbf{Implementation Details.} \quad
In the FluxGen pipeline, we generate images at a resolution of $1024\times 768$, using the guidance scale of $4$ and timesteps of $20$. In all experiments, we exclusively use 350,000 images generated by FluxGen, with no external data involved. We randomly center-crop the images to the sizes of $1024\times 768$, $768\times 768$, and $512\times 768$ to accommodate different aspect ratios. We train our FluxIR model using the AdamW~\cite{loshchilov2017decoupled} optimizer with a learning rate of $1\times 10^{-5}$. The training process takes about $3.5$ days on $4$ NVIDIA H800 GPUs with a batch size of $64$. For inference, we adopt $20$ sampling steps to generate our IR results across all of our experiments.

\noindent\textbf{Metrics.} \quad
Following  \cite{yu2024scaling}, we use PSNR, SSIM, and LPIPS as full-reference metrics and MANIQA, CLIPIQA, and MUSIQ as non-reference metrics\footnote{We utilize the GitHub repository - \url{https://github.com/chaofengc/IQA-PyTorch} to evaluate non-reference metric scores with model names ``maniqa-pipal'', ``clipiqa+\_vitL14\_512'', and ``musiq'', respectively.} for comprehensive evaluation. Our method, like other generative IR approaches~\cite{wang2024exploiting, lin2023diffbir, wu2024seesr, yu2024scaling, ai2024dreamclear}, achieves strong results in non-reference metrics but performs less competitively on full-reference metrics.

\subsection{Comparison with Existing Methods}
We conduct both quantitative and qualitative comparisons of our FluxIR with state-of-the-art image restoration methods, including early approaches (BSRGAN~\cite{zhang2021designing}, RealESRGAN\cite{wang2021real}, SwinIR~\cite{liang2021swinir} and DASR\cite{liang2022efficient}) and recent generative IR models (StableSR~\cite{wang2024exploiting}, DiffBIR~\cite{lin2023diffbir}, ResShift~\cite{yue2024resshift}, SinSR~\cite{wang2024sinsr}, SeeSR~\cite{wu2024seesr}, SUPIR~\cite{yu2024scaling}, DreamClear~\cite{ai2024dreamclear}).

\noindent\textbf{Quantitative Comparisons.} \quad
\cref{tab:sota} shows the quantitative comparisons with the existing SOTAs on the four datasets. 
Our FluxIR model achieves the best score in MANIQA with a significant margin across all synthetic and real-world datasets. On the \textit{RealLQ250} dataset, our method significantly outperforms other methods across all non-reference metrics, while MANIQA even surpasses the second-best method by over 6.53\%. On the other three synthetic and real-world datasets, our approach achieves the best or second-best scores in all non-reference metrics (CILPIQA, MUSIQ, and MANIQA). It is worth mentioning that our proposed method trained within $14$ GPU-days significantly reduces training costs by approximately $93.5$\% and $91.5$\% compared to SUPIR~\cite{yu2024scaling} and DreamClear~\cite{ai2024dreamclear}, respectively. 

\noindent\textbf{Qualitative Comparisons.} \quad
We present a qualitative comparison on the synthetic dataset \textit{DIV2K-Val} in \cref{fig:visual_synthetic}. Our method demonstrates a notable advantage in generating high-frequency details, such as lip texture, hair, eyelashes, and pupils in portrait images. In contrast, SUPIR~\cite{yu2024scaling} and DreamClear~\cite{ai2024dreamclear} fail to accurately reproduce human skin textures, and SeeSR~\cite{wu2024seesr} produces overly smooth results, lacking the natural skin texture and fine details.
For the real-world dataset \textit{RealLQ250}, we provide a qualitative comparison in \cref{fig:visual_real}. In the first row, which shows restoration results for a low-quality image of a cat, our method effectively reconstructs realistic high-frequency details such as fur, whiskers, and eyes. SeeSR~\cite{wu2024seesr}  restores some fur and whiskers but lacks overall realism, while other methods fail to produce a high-quality, realistic result. In the second row, focusing on text recovery, our method achieves the most accurate text reconstruction, whereas other methods introduce various text distortions. Specifically, SUPIR~\cite{yu2024scaling} produces blurry results, and SeeSR~\cite{wu2024seesr} mistakenly recovers incorrect text content.

\subsection{Ablation Study}
\noindent\textbf{Effectiveness of SE layer.} \quad
We evaluate the impact of different ranks of our SE layer by setting $r$ to $16$, $32$, $64$, and $128$, respectively. Another variant is replacing the SE layer with a single MLP, \ie a full rank setting. As shown in \cref{tab:ablation_rank}, the SE layer with $r = 32$ achieves the best performance on CLIPIQA, MANIQA, and CLIPIQA.
Compared to using full-rank MLPs, the SE layer with low-rank results in a comparable performance while saving over 72\% of the parameters.  This demonstrates the effectiveness of our designed light-weighted SE layers in controlling the Flux model. By utilizing our SE layers with a rank of $32$, we can outperform other state-of-the-art (SOTA) methods using only 0.4B adapter parameters, which is only 30.8\% and 18.6\% the size of SUPIR and DreamClear, respectively.

\begin{table}[t]
    \captionsetup{font=small}
    \small
    \centering
    \renewcommand\arraystretch{1.0}
    \setlength\tabcolsep{4pt}
    \caption{Ablation results of various SE layer ranks on the \textit{RealLQ250} dataset. The number of parameters in FluxIR without the SE layers is 387M.}
    \vspace{-7pt}
    \begin{tabular}{lcccl}
        \toprule
        {\footnotesize Rank} & {\footnotesize CLIPIQA $\uparrow$} & {\footnotesize MUSIQ $\uparrow$} &  {\footnotesize MANIQA $\uparrow$} & \# {\footnotesize Params } \\
        \midrule
        16 & 0.5389 & 69.72 & 0.6222 & 387M + 11.6M \\
        32 & 0.5639 & 70.78 & 0.6314 & 387M + 22.8M \\
        64 & 0.5529 & 70.37 & 0.6306 & 387M + 45.2M \\
        128 & 0.5606 & 70.54 & 0.6292 & 387M + 90.0M \\
        Full & 0.5720 & 70.79 & 0.6324 & 387M + 1076.2M \\
        \bottomrule
    \end{tabular}
    \label{tab:ablation_rank}
    \vspace{-5pt}
\end{table}
\begin{figure}[t]
    \centering
    \includegraphics[width=1\linewidth]{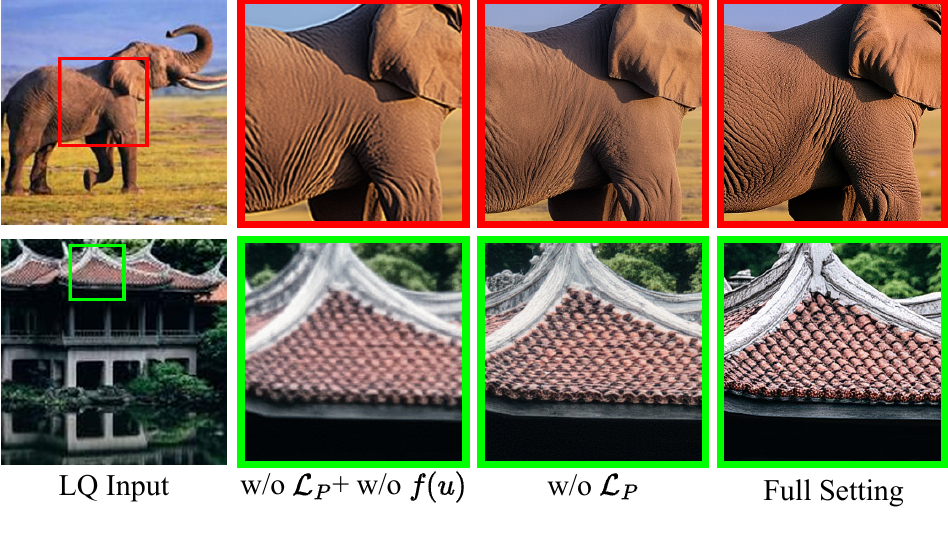}
    \vspace{-20pt}
    \caption{
        The visual comparison results of our training strategies: timestep sampling function $f(u)$ and $\mathcal{L}_{P}$ loss. Please zoom in for a better view.
    }
    \label{fig:ablation_optimization}
    \vspace{-15pt}
\end{figure}
\noindent\textbf{Effectiveness of Training Strategies.} \quad
We conduct ablation studies to evaluate our proposed training strategies: timestep sampling function $f(u)$ and $\mathcal{L}_{P}$ loss. 
Firstly, we use the logit-normal sampling strategy~\cite{esser2024scaling} to train our model without $\mathcal{L}_{P}$ loss as a baseline. Then we include our proposed timestep sampling function $f(u)$ as the second setting.
We compare these two experiments with our full training strategies in the \cref{tab:ablation_optimization}, which clearly demonstrates the significant performance improvement of the new sampling function and $\mathcal{L}_{P}$ loss. \cref{fig:ablation_optimization} illustrates that the logit-normal sampling produces low-quality content, while our proposed sampling function $f(u)$ effectively addresses this issue by bridging the gap between training and inference. Additionally, applying the pixel-space loss $\mathcal{L}_{P}$ allows for the restoration of more high-frequency details.
\begin{table}[h]
    \captionsetup{font=small}
    \small
    \centering
    \renewcommand\arraystretch{1.1}
    \setlength\tabcolsep{4pt}
    \caption{Ablation comparison results of our training strategies on the \textit{RealLQ250} dataset. }
    \vspace{-10pt}
    \begin{tabular}{lccc}
        \toprule
        Strategy & CLIPIQA $\uparrow$ & MUSIQ $\uparrow$ & MANIQA $\uparrow$ \\
        \midrule
        w/o $\mathcal{L}_{P}$ and w/o $f(u)$ & 0.4416 & 62.18 & 0.5720 \\
        w/o $\mathcal{L}_{P}$ & 0.5082 & 68.53 & 0.6128 \\
        Full Setting & \textbf{0.5639} & \textbf{70.78} & \textbf{0.6314} \\
        \bottomrule
    \end{tabular}
    \label{tab:ablation_optimization}
    \vspace{-5pt}
\end{table}

\begin{figure}[t]
    \centering
    \includegraphics[width=1\linewidth]{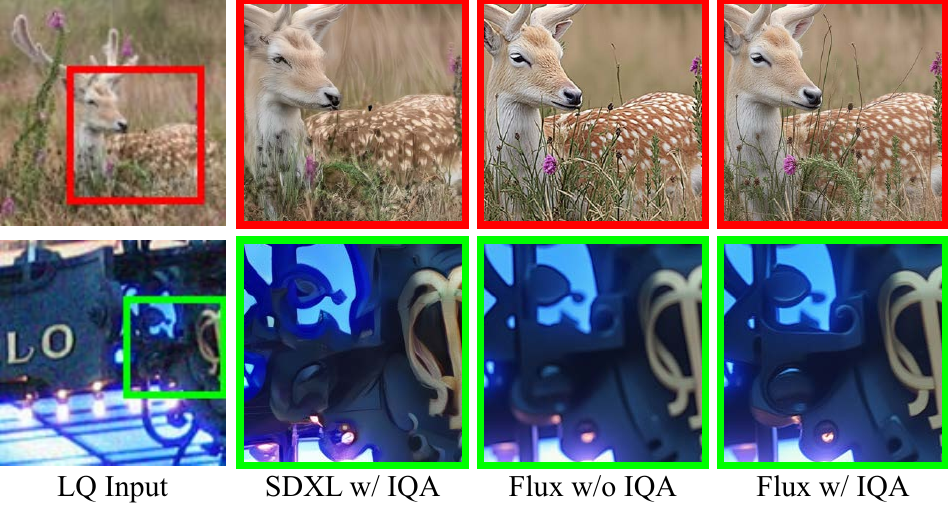}
    \vspace{-20pt}
    \caption{
        The visual comparisons of different FluxGen settings, where we study different T2I models, \ie SDXL and Flux, and the usage of IQA selections. Please zoom in for a better view.
    }
    \label{fig:ablation_SDXL}
    \vspace{-10pt}
\end{figure}

\noindent\textbf{Effectiveness of FluxGen.} \quad
We follow the same settings of our FluxGen pipeline but replace the T2I model from Flux with SDXL to generate $10,000$ images. Additionally, we remove the IQA-based selection from the FluxGen pipeline.  The results in \cref{tab:ablation_T2I} demonstrate that a high-quality image dataset from the Flux model, combined with IQA selection, leads to high-quality IR results. \cref{fig:ablation_SDXL} shows that the visual results based on the SDXL dataset are abysmal, exhibiting a smearing and distortion effect.  In contrast, images generated by Flux deliver the best visual aesthetic quality. Our IQA selection can eliminate poorly generated images to achieve more realistic results.  The samples of generated images by FluxGen can be found in the supplemental materials. 

\begin{table}[h]
    \vspace{-5pt}
    \captionsetup{font=small}
    \small
    \centering
    \renewcommand\arraystretch{1.0}
    \setlength\tabcolsep{4pt}
    \caption{Ablation results of different datasets generated by three FluxGen settings on the \textit{RealLQ250} dataset.}
    \vspace{-7pt}
    \begin{tabular}{lccc}
        \toprule
        FluxGen & CLIPIQA $\uparrow$ & MUSIQ $\uparrow$ &  MANIQA $\uparrow$ \\
        \midrule
        SDXL w/ IQA &  \textbf{0.5787} &  69.28 &  0.5630 \\
        Flux w/o IQA & 0.5254 & 68.66 & 0.6265 \\
        Flux w/ IQA &  0.5487 & \textbf{70.16} & \textbf{0.6267} \\
        \bottomrule
    \end{tabular}
    \label{tab:ablation_T2I}
    \vspace{-8pt}
\end{table}

\section{Conclusion}\label{sec: Conclusion}
In this paper, we squeezed out the powerful T2I model - Flux for image restoration, by acquiring training data from it and then building a lightweight adapter to control it.
To tackle the challenge of acquiring a large-scale high-quality image dataset, we proposed FluxGen for streamlined and highly efficient data generation.
We present FluxIR, a lightweight adapter designed to control the T2I model for real-world image restoration, where the SE layers broadcast the control signals to all Flux MM-DiT blocks and modulate both image and text embedding to enable precise and multi-modality control.
Extensive qualitative and quantitative experiments demonstrated that our method shows superior performance on image restoration tasks and significantly reduces the training cost of large-scale generative IR model, in comparison with previous state-of-the-arts. 
\newpage
{
    \small
    \bibliographystyle{ieeenat_fullname}
    \bibliography{main}
}

\newpage
\onecolumn
{\centering
\Large
\textbf{\thetitle}\\
\vspace{0.5em}Supplementary Material \\
\vspace{1.0em}
}
\section{More Ablation Studies.}
\label{sec:more_ablations} 

\begin{figure*}[th]
    \centering
    \includegraphics[width=1\linewidth]{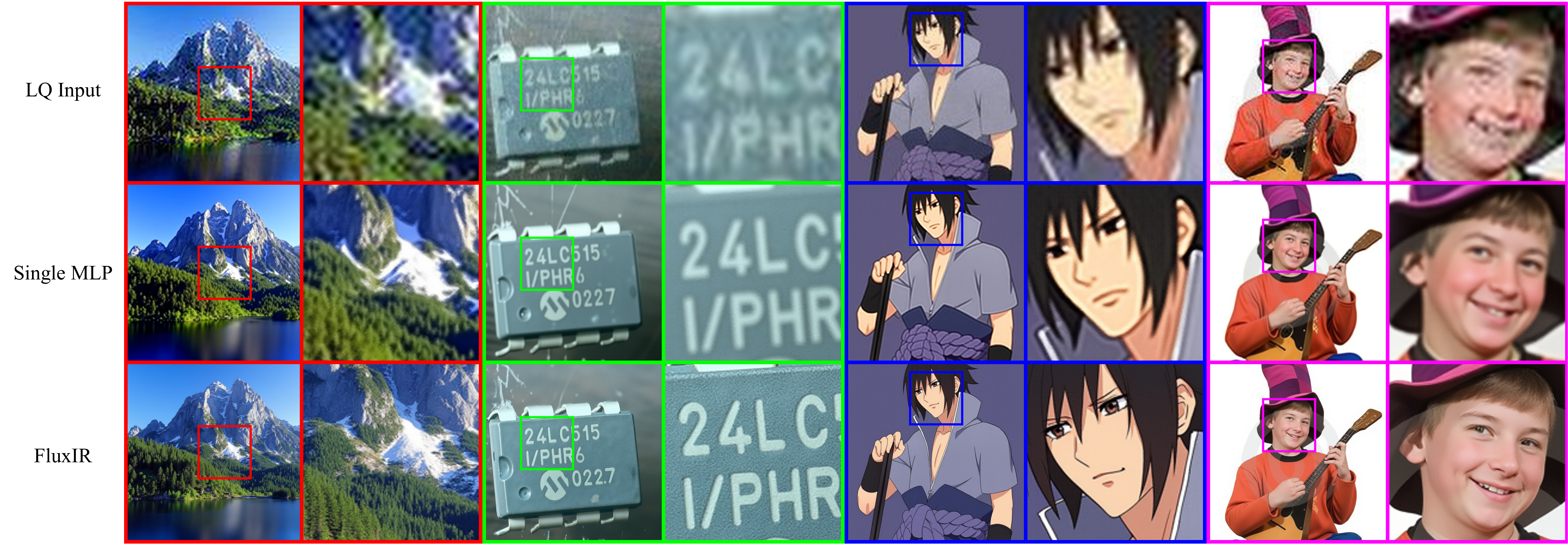}
    \caption{Visual results comparing the single MLP design and our multiple  SE layers. }
    \label{fig:ablation_MLP}
\end{figure*}

\paragraph{More Ablation of SE layers.}  \quad 
To validate the necessity of multiple SE layers in FluxIR, we replaced them with a single MLP, meaning a single full-rank MLP was used to control all Flux MM-DiT blocks simultaneously.
As shown in \cref{tab:ablation_MLP}, the single MLP will strongly degrade the generation performance in all metric scores. 
Additionally, \cref{fig:ablation_MLP} illustrates that using a single MLP limits the model's generative capacity, resulting in lower-quality outputs. These findings highlight that the optimal design for our Flux adapter is to provide dedicated control for each Flux MM-DiT block.

\begin{table}[htbp]
    \captionsetup{font=small}
    \small
    \centering
    \renewcommand\arraystretch{1.0}
    \setlength\tabcolsep{15pt}
    \caption{Comparison between Single MLP design and our multiple SE layers design on the \textit{RealLQ250} dataset.}
    \vspace{-5pt}
    \begin{tabular}{lccc}
        \toprule
        Control Layer & CLIPIQA $\uparrow$ & MUSIQ $\uparrow$ &  MANIQA $\uparrow$ \\
        \midrule
        FluxIR &  \textbf{0.5639} & \textbf{70.78} &  \textbf{0.6314} \\
        Single MLP & 0.5111 & 64.46 & 0.5547 \\
        \bottomrule
    \end{tabular}
    \label{tab:ablation_MLP}
\end{table}

\begin{figure*}[hbtp]
    \centering
    \includegraphics[width=0.57\linewidth]{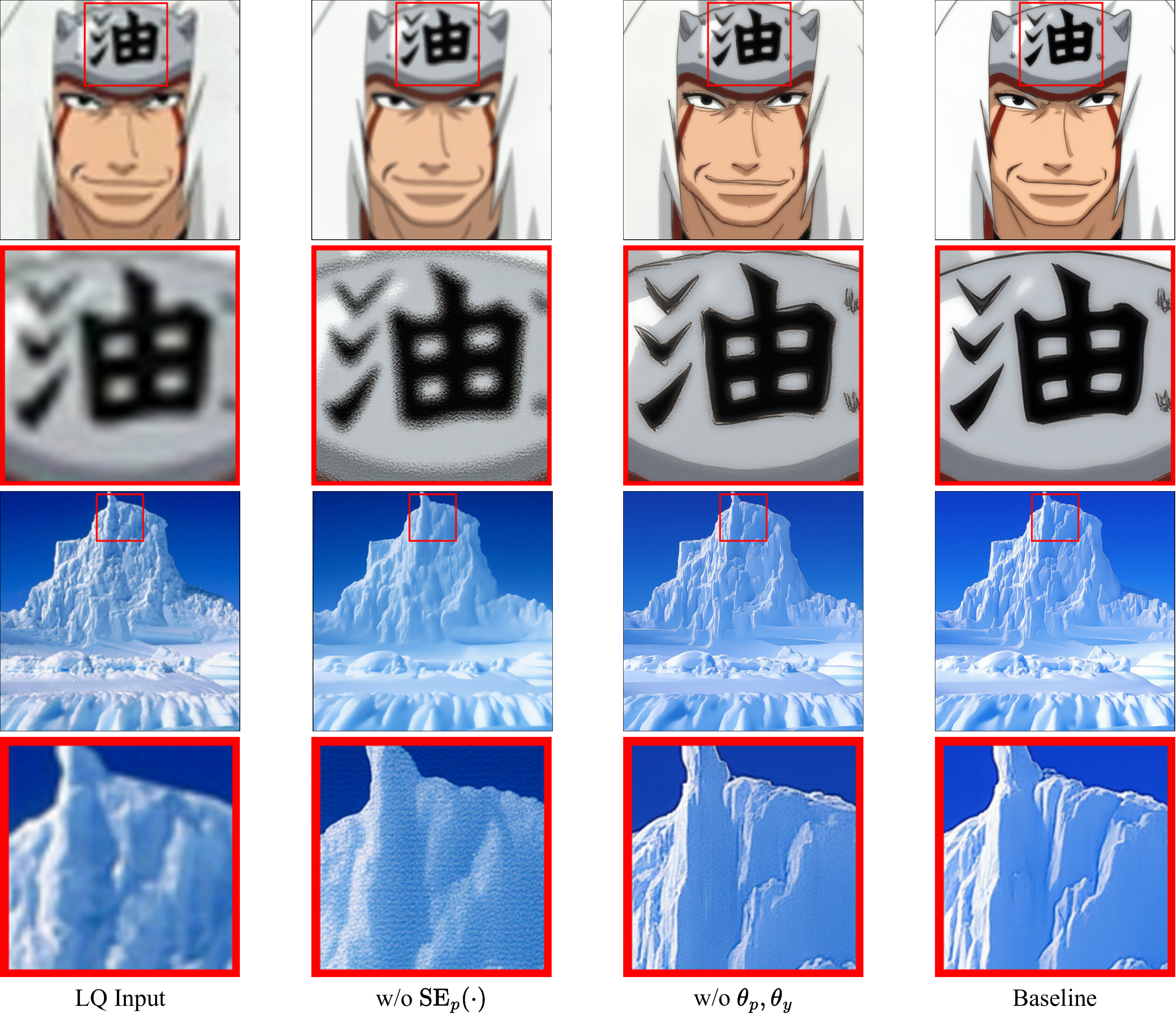}
    \caption{The visual comparisons of our multi-modality designs, \ie text branch SE layers $\text{SE}_p(\cdot)$ and trainable embeddings $\theta_p$, $\theta_y$. Please zoom in for a better view.}
    \label{fig:text_ablation}
\end{figure*}

\paragraph{Ablation on multi-modality designs.}  \quad 
In our proposed FluxIR, we introduce multi-modality controls on both the image and text information with a learnable T5~\cite{raffel2020exploring} embedding $\theta_p$ and a learnable CLIP~\cite{radford2021learning} embedding $\theta_y$. To justify the effectiveness of these designs, we evaluate the model variants by removing T5 embedding $\theta_y$ and CLIP embedding $\theta_y$ and text branch SE layer $\text{SE}_p(\cdot)$, respectively. 
\cref{tab:ablation_text} presents the quantitative results on \textit{RealLQ250} dataset. The results indicate that the text branch SE layer is crucial for enhancing the performance of our FluxIR model. The trainable T5 embedding $\theta_p$ and CLIP embedding $\theta_y$ show marginal differences from the baseline in evaluation metrics.  As shown in \cref{fig:text_ablation}, removing the text branch SE layers $\text{SE}_p(\cdot)$ leads to a significant decline in image restoration performance.
The trainable T5 embedding $\theta_p$ and CLIP embedding $\theta_y$ also contribute slight improvements in visual quality. The overall results demonstrate that the multi-modality design of FluxIR effectively boosts performance in the image restoration task.

\begin{table}[htbp]
    \captionsetup{font=small}
    \small
    \centering
    \renewcommand\arraystretch{1.0}
    \setlength\tabcolsep{15pt}
    \caption{Ablation results of multi-modality designs on the \textit{RealLQ250} dataset.}
    \vspace{-5pt}
    \begin{tabular}{lccc}
        \toprule
        Multi-Modality & CLIPIQA $\uparrow$ & MUSIQ $\uparrow$ &  MANIQA $\uparrow$ \\
        \midrule
        Baseline &  \textbf{0.5639} & \textbf{70.78} &  \textbf{0.6314} \\
        w/o $\theta_p$,$\theta_y$ & 0.5626 & 70.53 & 0.6308  \\
        w/o $\text{SE}_p(\cdot)$ &  0.5259 & 67.63 & 0.6266 \\

        \bottomrule
    \end{tabular}
    \label{tab:ablation_text}
\end{table}

\section{Samples of Training dataset built by FluxGen}
\begin{figure*}[th]
    \centering
    \includegraphics[width=1.\linewidth]{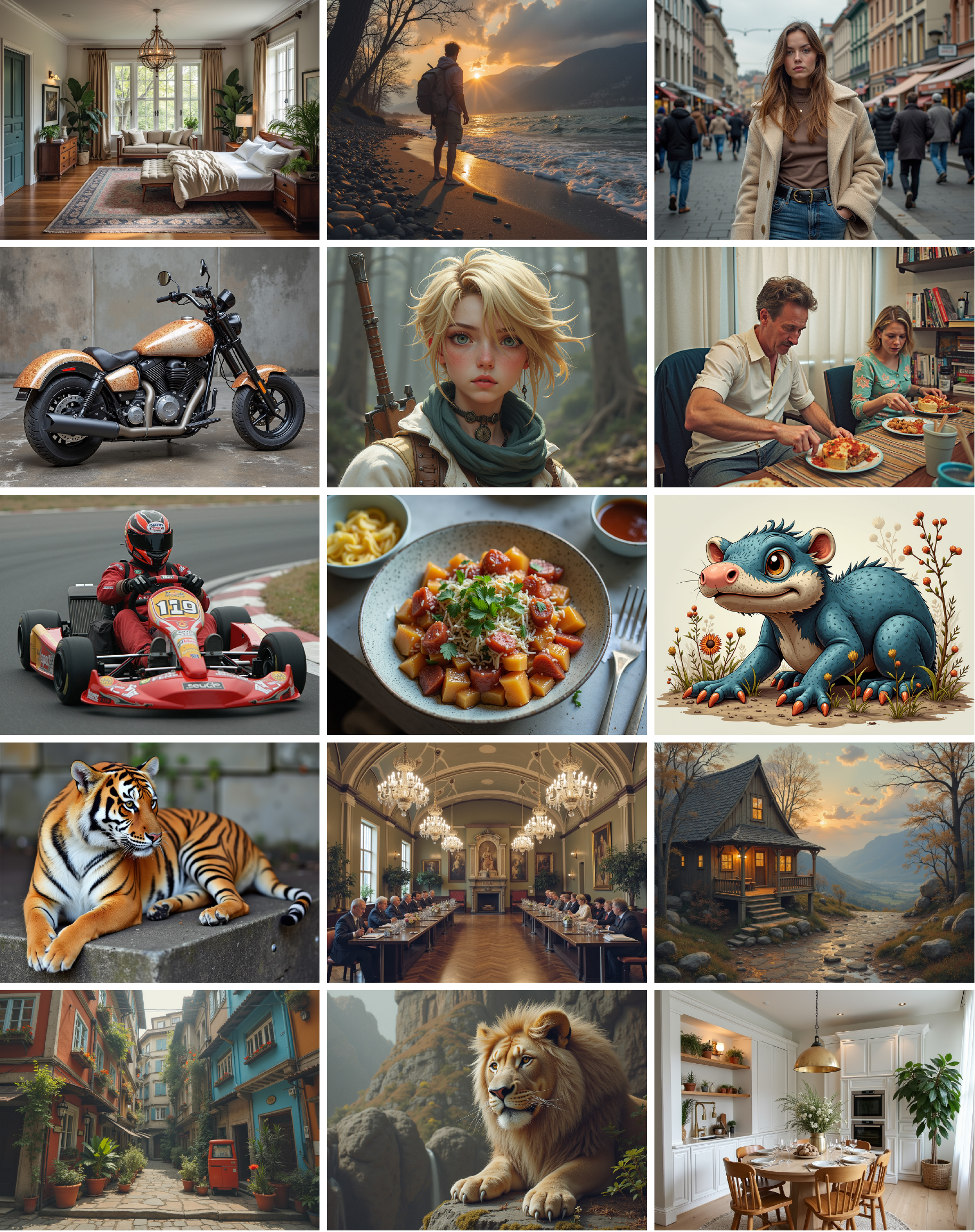}
    \caption{Samples of generated images by FluxGen pipeline.}
    \label{fig:fluxgen_flux}
\end{figure*}

\begin{figure*}[th]
    \centering
    \includegraphics[width=1\linewidth]{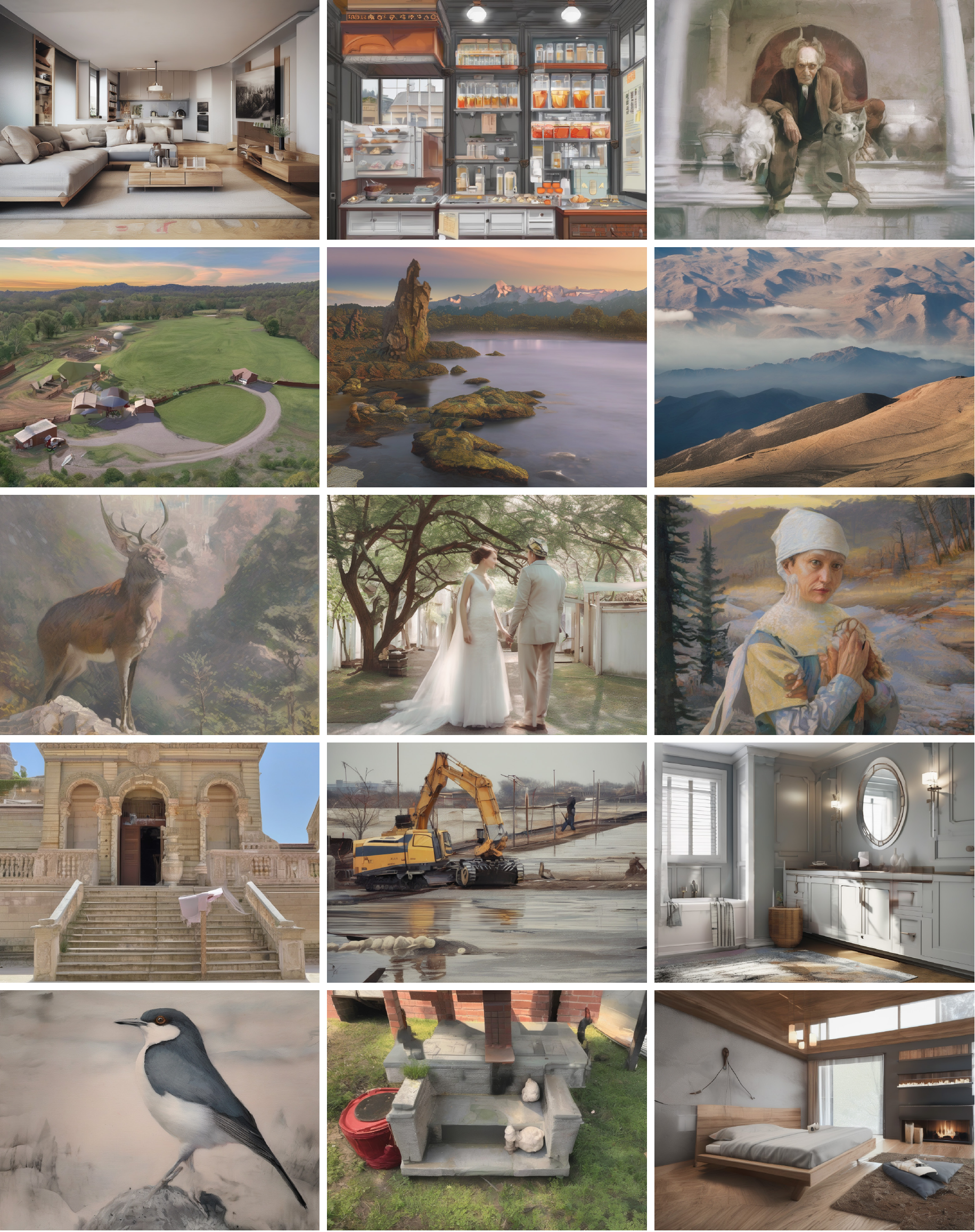}
    \caption{Samples generated by the SDXL.}
    \label{fig:fluxgen_sdxl}
\end{figure*}

\begin{figure*}[th]
    \centering
    \includegraphics[width=0.95\linewidth]{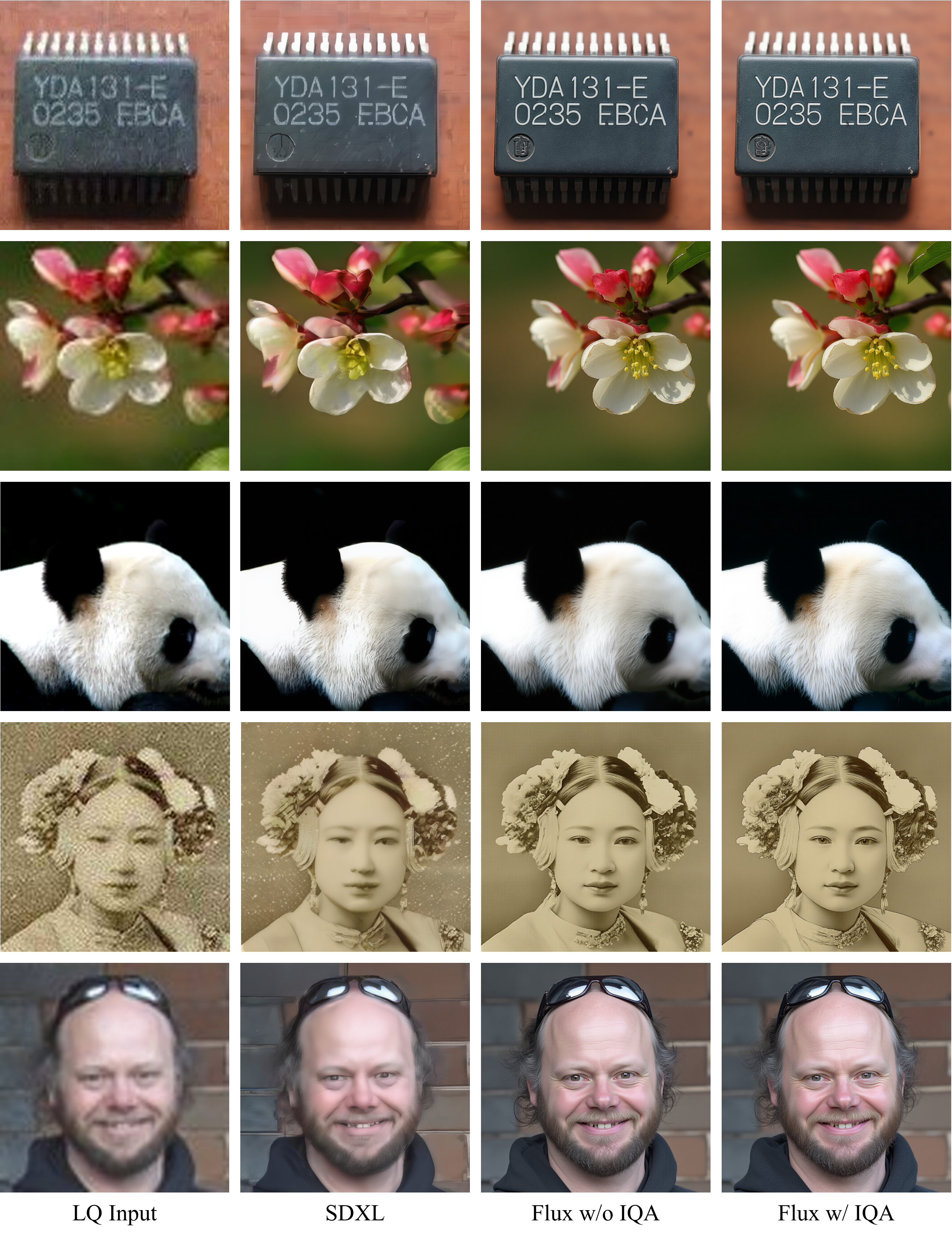}
    \caption{The visual comparisons of different FluxGen settings, where we study different T2I models, \ie SDXL and Flux, and the usage of IQA selections. Please zoom in for a better view.}
    \label{fig:tex_Supp_FluxGen_ablation}
\end{figure*}

In this section, we present the dozens of samples produced by the FluxGen pipeline with the resolution of $1,024\times 768$. \cref{fig:fluxgen_flux} illustrates our generated training dataset obtained with an empty prompt, and demonstrates that an empty prompt is sufficient to produce diverse scene images with high resolution and aesthetic quality including cars, portraits, anime characters, animals, plants, food, buildings, indoor settings, furniture, the sea, and sunsets. We found that some ground truth images from FluxGen contain bokeh effects, which can occasionally cause localized blurriness in the restored results. However, based on subjective evaluations across four test datasets, the impact is minimal and acceptable. Similar issues could also arise in real-world datasets if not properly cleaned. Meanwhile, we show SDXL generated data in \cref{fig:fluxgen_sdxl}, which is also employed in the ablation studies. Without carefully designed prompt, SDXL cannot produce high-quality images for image restoration tasks.  \cref{fig:tex_Supp_FluxGen_ablation} shows more visual comparisons to further justify the effectiveness of FluxGen on the choice of text-to-image model and IQA selection.
Furthermore, we generated 2,000 images from each of the five existing T2I models (PixelArt-$\Sigma$~\cite{chen2024pixart}, Sana~\cite{xie2024sana}, SDXL~\cite{podell2023sdxl}, Playground~\cite{li2024playground}, and Flux.1-dev~\cite{flux2024}) for evaluation. As shown in \cref{tab:ablation_T2I_IQA}, the images generated by our FluxGen pipeline achieved superior IQA scores.
\begin{table}[htbp]
    \captionsetup{font=small}
    \small
    \centering
    \caption{Comparisons with existing T2I generation methods.}
    \begin{tabular}{c|cccccc}
        \toprule
        Metric & PixArt-$\Sigma$ & Sana & SDXL & Playground & Flux.1-dev & Ours \\
        \midrule
        CLIPIQA $\uparrow$ & 0.4981 & 0.5135 & 0.5821 & 0.5822 & 0.6763 & \textbf{0.7295} \\
        MUSIQ $\uparrow$ & 64.93 & 66.75 & 70.49 & 70.35 & 75.02 & \textbf{75.37} \\
        MANIQA $\uparrow$ & 0.5519 & 0.5944 & 0.5831 & 0.6666 & 0.6590 & \textbf{0.6962} \\
        \bottomrule
    \end{tabular}
    \label{tab:ablation_T2I_IQA}
\end{table}


\section{More Visualization Comparison.}
\begin{figure*}[th]
    \centering
    \includegraphics[width=0.95\linewidth]{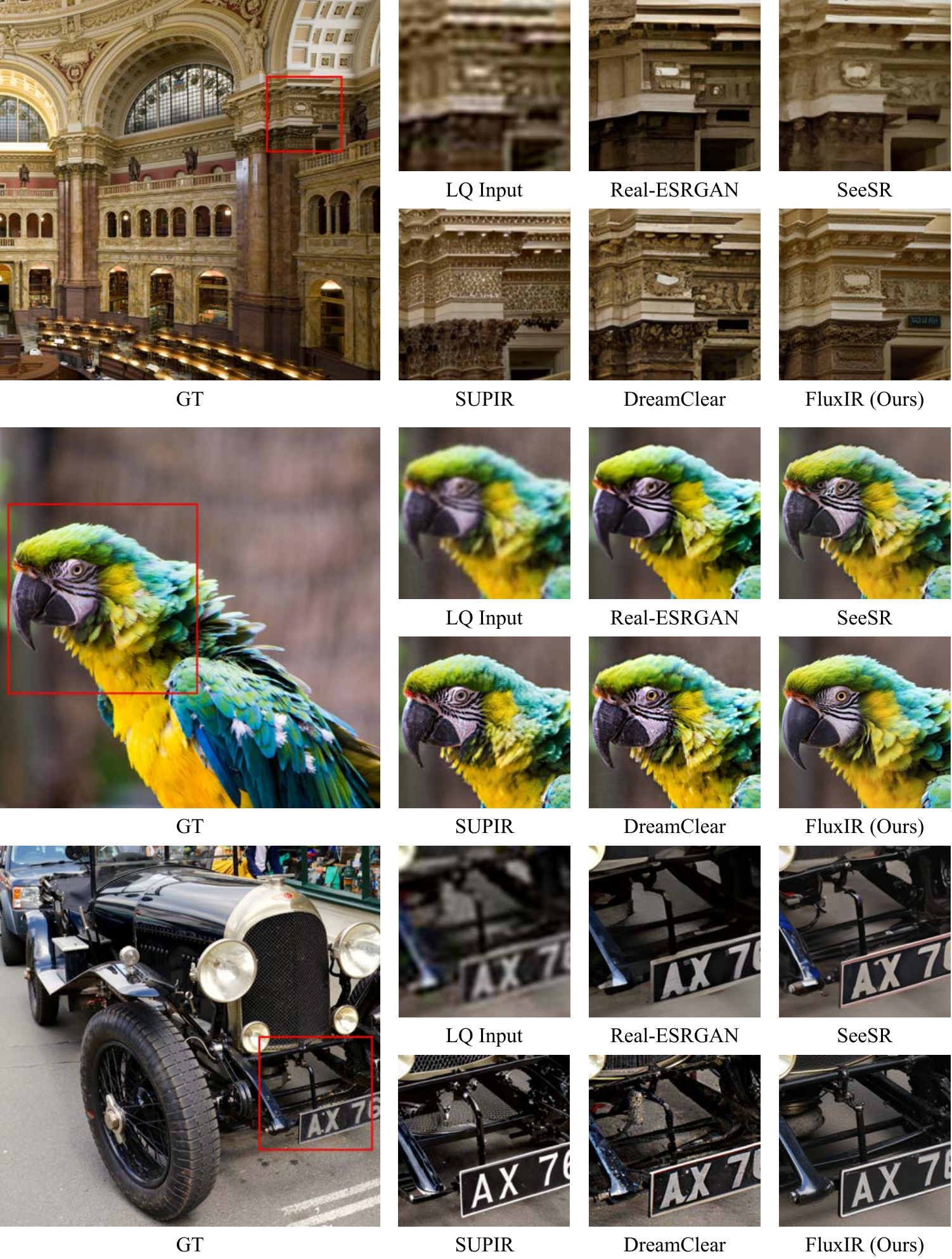}
    \caption{Visual comparison with SOTAs on \textit{DIV2K-Val} dataset. }
    \label{fig:comparison_div2k}
\end{figure*}

\begin{figure*}[th]
    \centering
    \includegraphics[width=0.95\linewidth]{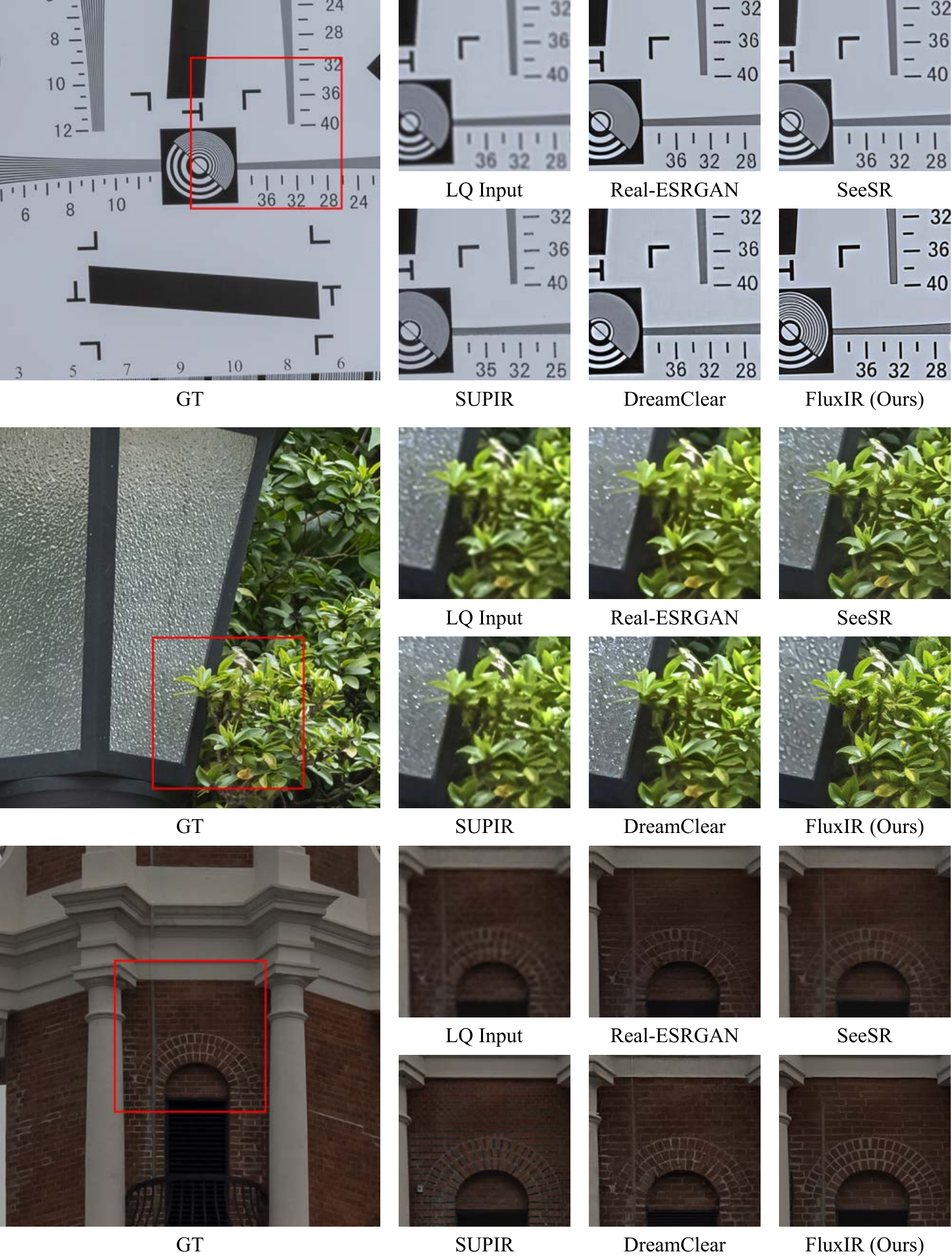}
    \caption{Visual comparison with SOTAs on \textit{RealSR} dataset. }
    \label{fig:comparison_realsr}
\end{figure*}

\begin{figure*}[th]
    \centering
    \includegraphics[width=0.95\linewidth]{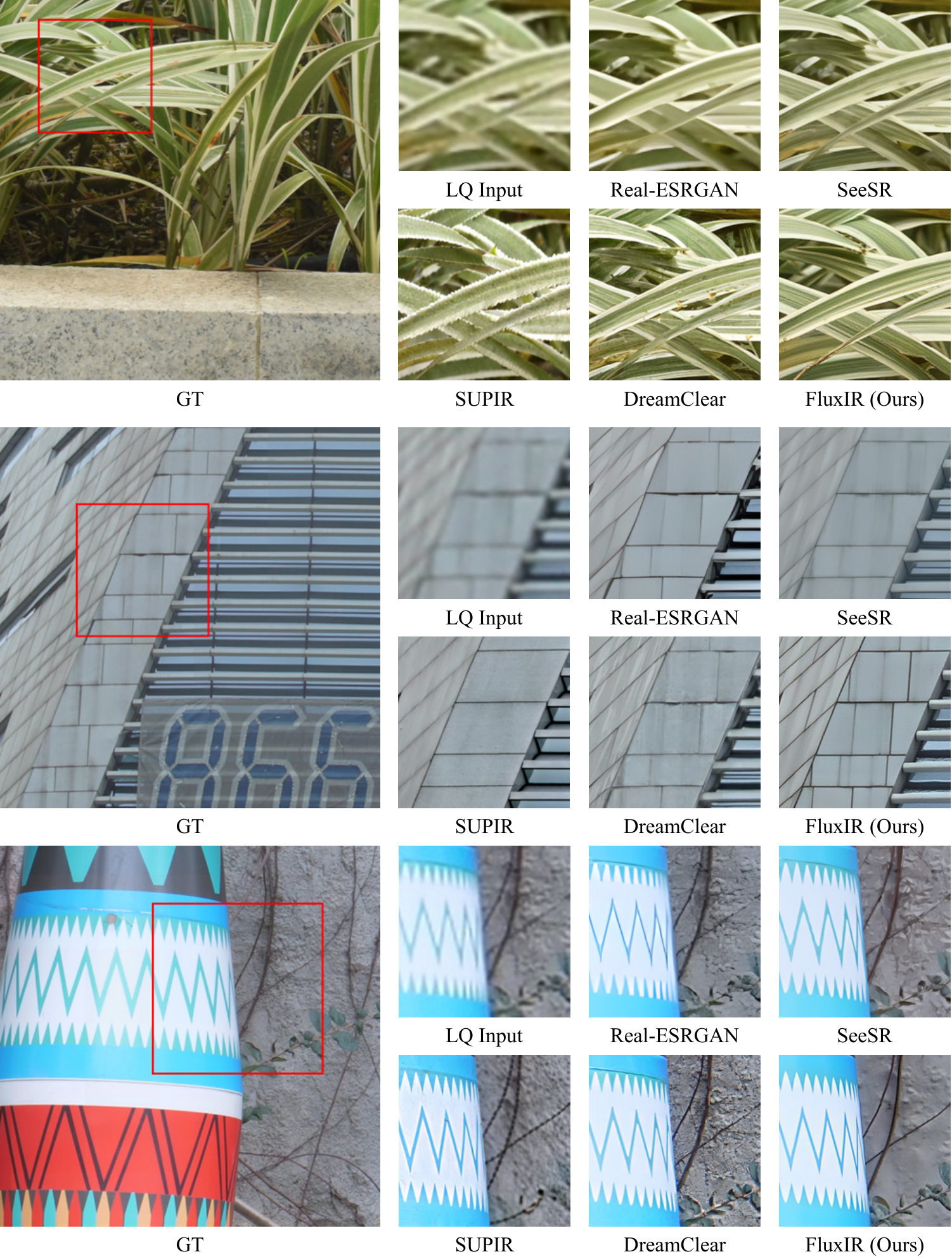}
    \caption{Visual comparison with SOTAs on \textit{DrealSR} dataset. }
    \label{fig:comparison_drealsr}
\end{figure*}

\begin{figure*}[th]
    \centering
    \includegraphics[width=0.95\linewidth]{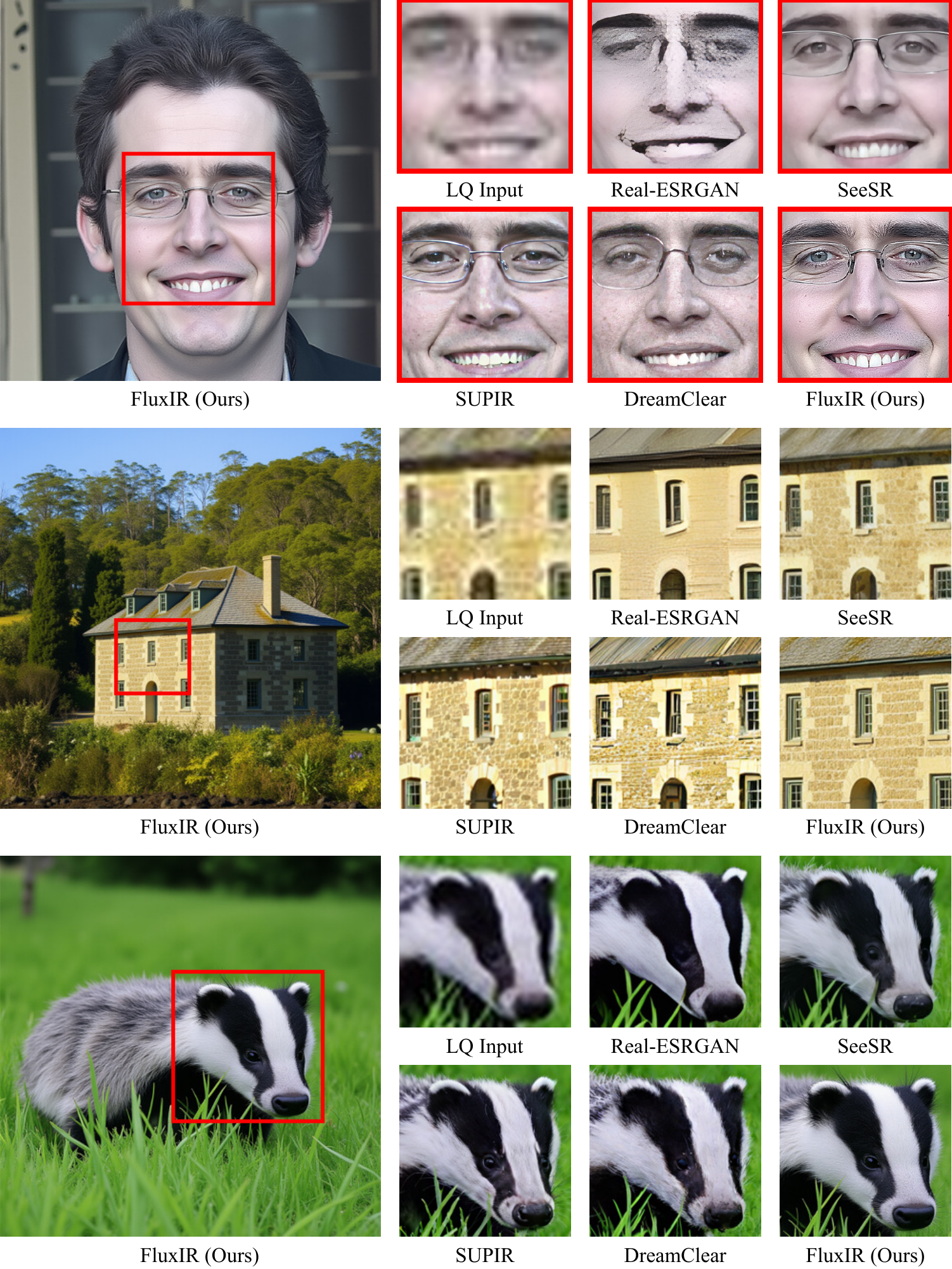}
    \caption{Visual comparison with SOTAs on \textit{RealLQ250} dataset. }
    \label{fig:comparison_reallq250}
\end{figure*}

Here, we provide additional visual results on synthetic and real-world datasets compared with state-of-the-art methods.  \cref{fig:comparison_div2k} presents the visual results on the \textit{DIV2K-Val}~\cite{agustsson2017ntire} dataset. \cref{fig:comparison_realsr} presents the visual results on the \textit{RealSR}~\cite{cai2019toward} dataset. \cref{fig:comparison_drealsr} presents the visual results on the \textit{DrealSR}~\cite{wei2020component} dataset. \cref{fig:comparison_reallq250} presents the visual results on the \textit{RealLQ250}~\cite{ai2024dreamclear} dataset. Our FluxIR achieves the best performance in terms of generation quality, texture details, and aesthetic quality. Please zoom in for a better view.


\newpage
{
    \small
    \bibliographystyle{ieeenat_fullname}
    \bibliography{main}
}


\end{document}